\PassOptionsToPackage{table,xcdraw}{xcolor}

\documentclass[sigconf]{acmart}

\AtBeginDocument{%
  }

\setcopyright{acmlicensed}
\copyrightyear{2018}
\acmYear{2018}
\acmDOI{XXXXXXX.XXXXXXX}
\acmConference[Conference acronym 'XX]{Make sure to enter the correct
  conference title from your rights confirmation email}{June 03--05,
  2018}{Woodstock, NY}

\acmISBN{978-1-4503-XXXX-X/2018/06}

\usepackage{overpic}

\usepackage{amssymb}

\usepackage{multirow}
\usepackage{subcaption}
\usepackage{graphicx}
\usepackage{algorithm}
\usepackage{algorithmic}

\usepackage[table]{xcolor}
\usepackage{colortbl}

\usepackage{pifont}

\usepackage[capitalize]{cleveref}
\crefname{section}{Sec.}{Secs.}
\Crefname{section}{Section}{Sections}
\Crefname{table}{Table}{Tables}
\crefname{table}{Table}{Tables}

\begin{document}

\title{DyST-XL: Dynamic Layout Planning and Content Control \\for Compositional Text-to-Video Generation}

\author{Weijie He$^{\ast}$}
\affiliation{%
\institution{Zhejiang University}
\city{}
\country{}}

\author{Mushui Liu$^{\ast \dagger}$}
\affiliation{%
\institution{Zhejiang University}
\city{}
\country{}}

\author{Yunlong Yu$^\ddagger$}
\affiliation{%
\institution{Zhejiang University}
\city{}
\country{}}

\author{Zhao Wang}
\affiliation{%
\institution{Zhejiang University}
\city{}
\country{}}

\author{Chao Wu}
\affiliation{%
\institution{Zhejiang University}
\city{}
\country{}}

\thanks{$\ast$ Equal Contribution. $\dagger$ Project Leader. $^\ddagger$ Corresponding author.}

\begin{abstract}
Compositional text-to-video generation, which requires synthesizing dynamic scenes with multiple interacting entities and precise spatial-temporal relationships, remains a critical challenge for diffusion-based models. Existing methods struggle with layout discontinuity, entity identity drift, and implausible interaction dynamics due to unconstrained cross-attention mechanisms and inadequate physics-aware reasoning.
To address these limitations, we propose DyST-XL, a \textbf{training-free} framework that enhances off-the-shelf text-to-video models (e.g., CogVideoX-5B) through frame-aware control. DyST-XL integrates three key innovations: (1) A Dynamic Layout Planner that leverages large language models (LLMs) to parse input prompts into entity-attribute graphs and generates physics-aware keyframe layouts, with intermediate frames interpolated via trajectory optimization; (2) A Dual-Prompt Controlled Attention Mechanism that enforces localized text-video alignment through frame-aware attention masking, achieving precise control over individual entities; and (3) An Entity-Consistency Constraint strategy that propagates first-frame feature embeddings to subsequent frames during denoising, preserving object identity without manual annotation.
Experiments demonstrate that DyST-XL excels in compositional text-to-video generation, significantly improving performance on complex prompts and bridging a crucial gap in training-free video synthesis. The code is released in \url{https://github.com/XiaoBuL/DyST-XL}.
\end{abstract}



\begin{teaserfigure}
  \centering
  \includegraphics[width=0.95\linewidth]{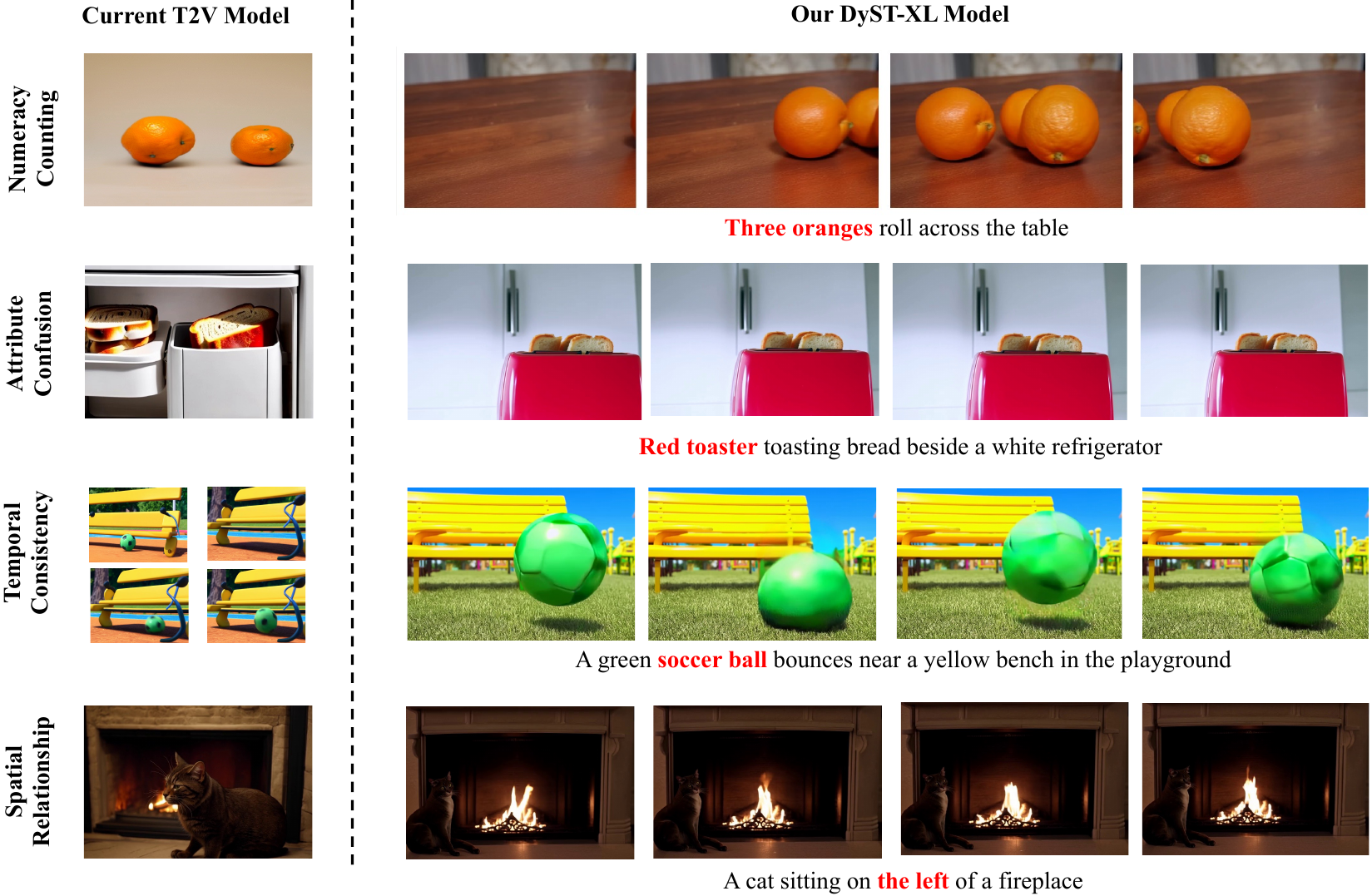}
  \caption{Common failure cases in current T2V generation: (1) Numeracy counting (three oranges), (2) Attribute Confusion (red toaster), (3) Temporal Consistency (missing soccer ball in the second frame), and (4) Spatial Relationship (left). Our proposed DyST-XL addresses these issues through dynamic layout planning and content control.}
   \vspace{1mm}
  \label{fig:teaser}
\end{teaserfigure}


\maketitle

\section{Introduction} \label{sec:introduction}

Diffusion models \cite{ho2020ddpm, ddim, stable_diffusion} have revolutionized conditional content generation, powering groundbreaking advances in both text-to-image \cite{sd3,flux} and text-to-video generation \cite{cogvideox,stepvideot2v}. The introduction of transformer-based architectures like Diffusion Transformers (DiT) \cite{DiT} has further elevated this paradigm through scale-aware designs that achieve spatial-temporal coherence in generated outputs. However, generating semantically consistent visual content from compositional texts that specify multiple interacting entities, their attributes, and dynamic spatial relationships remains a critical challenge in cross-modal synthesis, exposing the limitations in existing models' hierarchical reasoning capabilities.

The existing approaches for compositional generation primarily fall into two paradigms: \textbf{control-based conditioning} and \textbf{training-free semantic parsing}. Control-based techniques like ControlNet \cite{controlnet} enhance generation fidelity through explicit conditional guidance, yet their reliance on manually annotated, task-specific control signals inherently limits cross-domain adaptability, as each new domain demands specialized training data and model variants. In contrast, training-free methods like Regional Prompting FLUX \cite{regionalpromptflux} and RPG \cite{RPG} bypass supervision by leveraging large language models (LLMs) \cite{zhu2023minigpt, gpt-4} to parse multi-entity prompts via chain-of-thought reasoning—decomposing text into entity-attribute pairs and spatially segmenting images into non-overlapping regions.

While remarkable progress has been made in text-to-image generation, extending these capabilities to compositional text-to-video generation introduces additional challenges in modeling dynamic evolution. \textbf{(i) Layout Discontinuity:} The absence of physics-aware interpolation between keyframe layouts leads to non-smooth transitions, manifesting as abrupt object repositioning or scaling artifacts across consecutive frames. \textbf{(ii) Cross-Frame Entity Drift:} Unconstrained feature diffusion in temporal denoising processes causes progressive divergence in object appearance, violating identity preservation constraints essential for coherent storytelling. \textbf{(iii) Interaction Dynamics Misalignment:} Physically implausible motions emerge when synthesizing multi-entity interactions due to unconstrained cross-attention mechanisms. Current post-hoc corrections lack a unified framework for joint text-to-layout-to-motion optimization. Some failure cases are shown in \cref{fig:teaser}.

In this paper, we present \textbf{DyST-XL}, a training-free framework that improves compositional video generation capabilities in existing large-scale text-to-video models (CogVideoX-5B \cite{cogvideox} is used in this work) through three key components, i.e., a Dynamic Layout Planner for structured scene decomposition, a Dual-prompt Controlled Attention Mechanism to precise control over both global and entity-level content generation, and an Entity-consistency Constraint strategy that enforces temporal stability across object attributes and interactions.

The first component addresses ``\textbf{where to generate}" by orchestrating entity trajectories via LLM-guided diffusion optimization. Leveraging chain-of-thought reasoning, a large language model (LLM) parses input prompts to extract entities, their mobility attributes, and spatial relationships, generating coherent initial and final layouts for movable objects. Intermediate layouts are obtained via linear interpolation, ensuring smooth motion transitions across frames. 

The last two components collectively govern ``\textbf{how to generate}" by modulating spatiotemporal attention. First, a frame-aware attention-masking strategy restricts text-video interactions to localized regions, enforcing precise alignment between textual attributes and visual elements while mitigating implausible cross-frame interference. Second, to maintain entity identity consistency, we extract feature embeddings from the bounding box that defined entities in the first frame and iteratively fuse the reference representations into the corresponding regions of subsequent frames during denoising steps. This feature propagation strengthens temporal coherence without auxiliary optimization or manual annotation.

By integrating layout planning with spatiotemporal attention modulation, DyST-XL enables complex video generation that handles overlapping entities, dynamic motions, and multi-attribute specifications without user input. Notably, our approach operates without model retraining or handcrafted control conditions, offering a scalable solution for conditional video synthesis. Experiments demonstrate that our method achieves state-of-the-art performance on compositional prompts, bridging a critical gap in \textbf{training-free} text-to-video generation.

\section{Related Work} \label{sec:related}

\subsection{Text-to-Video Generation}
Text-to-video generation methods can be broadly categorized into two paradigms: (1) text-to-image-based approaches, which extend the pre-trained image diffusion models with temporal modeling, and (2) end-to-end video synthesis frameworks, designed from scratch to directly generate videos from text. 

Recent breakthroughs in text-to-image (T2I) diffusion models have catalyzed advances in text-to-video (T2V) synthesis, which can be broadly categorized into two paradigms. The first adapts pre-trained T2I models for video generation by augmenting them with temporal modeling capabilities. Methods such as \cite{tuneavideo}, \cite{animatediff}, and \cite{text2videozero} insert lightweight temporal attention or convolution layers into frozen T2I architectures, enabling video synthesis while preserving their spatial generative priors. These approaches minimize training costs by reusing pre-trained weights and focusing updates on temporal dynamics. The second paradigm involves video-native architectures fine-tuned end-to-end, such as those in \cite{videocrafter2, LVDM, cogvideox, mochi}, which explicitly model spatiotemporal relationships through dedicated designs—yielding high fidelity but requiring intensive computational resources and large-scale video datasets. The architectures of the existing end-to-end text-to-video generation primarily follow two paradigms: U-Net-based \cite{ronneberger2015u} and Diffusion Transformer (DiT)-based \cite{DiT} frameworks. Early U-Net models \cite{videocrafter2, LVDM, modelscope} dominated the field, leveraging encoder-decoder hierarchies and skip connections to capture multi-scale spatial-temporal features. However, recent DiT-based approaches \cite{cogvideox, mochi, opensora, opensoraplan} have shown promising results, benefiting from superior scalability, global context modeling, and cross-modal fusion, advantages that translate to enhanced temporal coherence and visual fidelity.

Our method enhances the compositional generation capabilities of the DiT paradigm through two critical improvements: (1) explicit spatial layout planning for complex scene structures and (2) cross-temporal entity preservation to maintain object consistency. 

\subsection{Compositional Generation}

When handling compositional texts that involve multiple entities or attributes, generative models often face challenges such as omissions and semantic conflations. To tackle these issues, researchers have proposed various strategies across the image and video generation domains.

In image generation, existing methods can be categorized into two main approaches. (1) Conditional Injection Methods: GLIGEN \cite{gligen} introduces trainable modules to incorporate spatial constraints, while InstanceDiffusion \cite{InstanceDiffusion} enforces instance-level control via a UniFusion network. MIGC \cite{MIGC} mitigates attribute leakage using a divide-and-conquer strategy, along with shading-based isolation techniques. These methods require careful architectural redesign and module training. (2) Training-Free Paradigms: RPG \cite{RPG} introduces a training-free framework that leverages large language models (LLMs) to parse entities, plan layouts via chain-of-thought reasoning, and inject region-aware text prompts into diffusion features. This paradigm is extended to Diffusion Transformers (DiT) in Regional Prompt FLUX \cite{regionalpromptflux}, enabling spatial control via attention masking without the need for model retraining.

In the video generation domain, solutions are focused on two main directions: (1) Spatiotemporal Control: VideoTetris \cite{videotetris} employs spatiotemporal compositional diffusion to govern regional dynamics, while DreamRunner \cite{dreamrunner} extends self-attention into 3D spatial-temporal-region operations to ensure consistency across frames. (2) Training-Free Alternatives: VideoRepair \cite{videorepair} iteratively refines video outputs using multimodal LLM feedback. However, it faces challenges such as cascading inference latency due to its reliance on multiple models and instability arising from error accumulation during regeneration cycles.

In contrast to prior approaches reliant on manual region tuning, our method achieves adaptive object arrangement through foundation model-guided spatial planning, coupled with time-aware feature blending to ensure temporal coherence.

\section{Method}

\begin{figure*}
    \centering
    \includegraphics[width=1.0\linewidth]{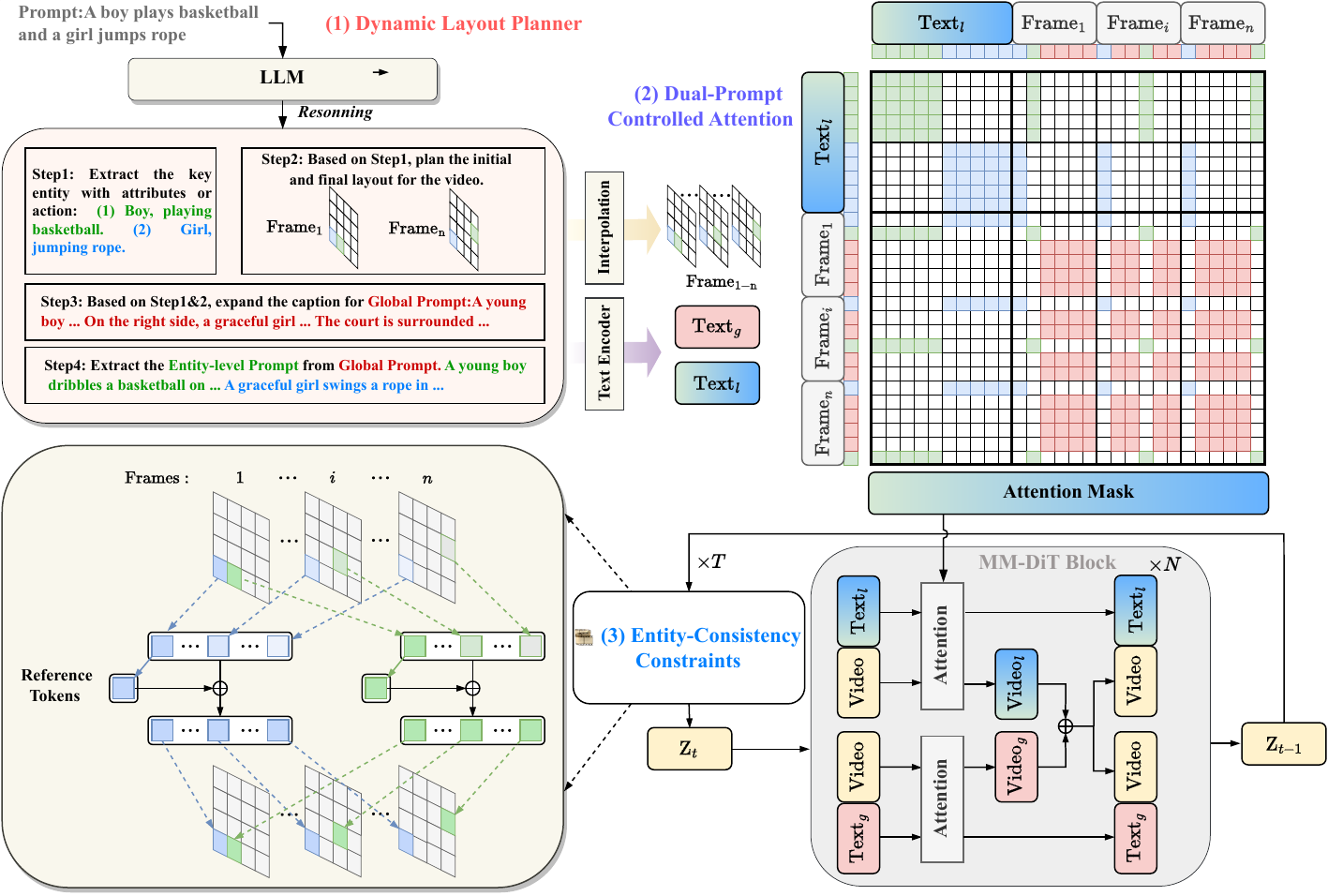}
    \caption{Illustration of our compositional text-to-video generation process. An LLM is used to parse the input prompt to extract entities and their attributes and generates global text prompts, entity-level text prompts, and dynamic bounding boxes suitable for the DiT text-to-video models. A dual-prompt controlled attention is used to control the video generation. During the denoising stage, an entity-consistency constraint strategy is used to ensure temporal consistency across frames. }
    \label{fig:enter-label}
\end{figure*}

In this work, we focus on the compositional text-to-video (T2V) generation, which requires synthesizing temporally coherent videos from textual descriptions involving multiple entities, dynamic interactions, and evolving spatial layouts.
Central to our approach is the adaptation of Diffusion Transformers (DiT)~\cite{DiT}, which treats the spatiotemporal latent representations as token sequences and processes them with textual tokens in the self-attention mechanism. We first briefly introduce the process of the existing T2V methods.

The generative process begins by progressively denoising a Gaussian latent tensor $\mathbf{z}_T \sim \mathcal{N}(0, I)$ into a structured latent video $\mathbf{z}_0 \in \mathbb{R}^{F \times H \times W \times D}$, where $F$,  $H$, $W$, and  $D$ denote the number of frames, height, width, and dimension respectively.
These tokens are reorganized into a sequence of length $(F \times H \times W) \times D$ to model spatiotemporal relationships. The final video $x_0$ is decoded from $z_0$ using a pre-trained decoder $\mathcal{D}$. At each timestep $t$, the denoising network $\epsilon_\theta$ predicts the noise component to iteratively refine the latent:
\begin{equation}
\mathbf{z}_{t-1} = f\left(\mathbf{z}_t, \epsilon_\theta(\mathbf{z}_t, C, t)\right),
\end{equation}
where $C$ represents the textual conditioning prompt, concatenated with the latent tokens, the function $f$ implements the diffusion update rule defined by the scheduler~\cite{ho2020ddpm,ddim,sd3}.
The denoising network $\epsilon_\theta$ consists of $N$ stacked DiT blocks~\cite{DiT}, each integrating multi-head self-attention and feedforward layers. To preserve positional coherence, spatial and temporal positional embeddings are injected either by direct addition to patch tokens or through query/key modulation in attention layers.

To address the critical challenges of layout discontinuity and cross-frame entity drift in compositional T2V generation, we propose a novel integration of three components into existing large-scale diffusion models (CogVideoX \cite{cogvideox} is used in this work): (1) a Dynamic Layout Planner for physics-aware motion trajectory design, (2) a Dual-Prompt Controlled Attention Mechanism that coordinates global scene context with localized attribute binding through spatiotemporal masking, and (3) an Entity-consistency Constraint strategy for identity-preserving synthesis across the frame generation. As illustrated in Fig.~\ref{fig:enter-label}, our framework systematically resolves spatial incoherence and temporal instability through synergistic layout planning and feature-guided denoising. We now detail these components.

\subsection{Dynamic Layout Planner}

While CogVideoX~\cite{cogvideox} achieves strong performance with verbose prompts, its domain-specific linguistic style training makes it prone to entity-attribute omission and semantic confusion \cite{vico} when processing compositional texts involving multiple interacting entities. To mitigate this, we employ a chain-of-thought prompting strategy that guides a large language model (LLM) to decompose complex textual inputs systematically. As shown in Fig.~\ref{fig:enter-label}, our approach extracts entities and their attributes and generates global text prompts, entity-level text prompts, and dynamic bounding boxes suitable for CogVideoX.

\textbf{Step 1: Entity-Attribute Parsing.}
The LLM parses the input text to identify entities (e.g., objects, characters) and their associated attributes (color, size), actions (e.g., ``running”), and spatial relationships (e.g., ``left of the tree”). This structured decomposition prevents semantic conflations in multi-entity scenarios.

\textbf{Step 2: Bounding Box Planning.}
The LLM categorizes entity motion dynamics, orchestrating bounding box trajectories: for static entities, it assigns fixed bounding boxes aligned with global spatial layouts, while for dynamic entities, it generates initial/final frame bounding boxes and computes intermediate regions through temporal linear interpolation, ensuring smooth spatiotemporal transitions (e.g., a car moving left-to-right while gradually shrinking in size). This process explicitly models positional continuity across frames using velocity profiles, avoiding abrupt spatial discontinuities in multi-attribute scenarios.

\textbf{Step 3: Global Prompt Generation.}
The LLM rewrites the original text prompt into a CogVideoX-compatible global prompt, preserving core semantics while adopting the model’s preferred linguistic style (e.g., verbose, entity-ordered descriptions). Spatial layouts from Step 2 are embedded implicitly to guide scene composition.

\textbf{Step 4: Entity-level Prompt Generation.}
Entity-specific descriptions (e.g., ``a young boy dribbles a basketball”) are extracted from the global prompt and concatenated into an entity-level prompt for regional feature injection.

The final output integrates the global prompt, entity-level prompt, and dynamic bounding boxes into a unified control schema, enabling precise attribute-layout-video alignment.

\subsection{Dual-Prompt Controlled Attention}

While global prompts align CogVideoX’s generation with its training distribution, they inadequately resolve entity-attribute confusion in compositional scenarios. To address this, we introduce local prompts paired with a spatiotemporal attention mask, enabling fine-grained control over text-video token interactions. The entity-level prompt (entity-specific descriptions) and global prompt (scene-level context) jointly condition generation through masked cross-attention, ensuring precise attribute-layout alignment.

\textbf{Attention Mask Design.} Following CogVideoX \cite{cogvideox}, our attention mechanism connects flattened video tokens $\mathbf{V} \in \mathbb{R}^{F \times H \times W \times D}$ with text tokens $\mathbf{T} \in \mathbb{R}^{L \times D}$, producing an attention matrix partitioned into four regions, i.e., T2T (Text-to-text interactions), T2V/V2T (Bidirectional text-video cross-attention), and V2V (Video token self-attention). Unlike CogVideoX \cite{cogvideox}, we substitute the text tokens in the original input prompt with those from extended global and entity-level prompts.

\textbf{T2T Masking.} The entity-level prompt comprises entity-specific descriptions partitioned into distinct \textbf{semantic classes}, where tokens describing the same entity (e.g., color, motion attributes of a red car) form a single class. We enforce \textbf{class-specific attention constraints}: each text token $T_i$ attends exclusively to tokens within its assigned class (e.g., all red car tokens), while attention between classes is masked. This prevents cross-entity attribute leakage (e.g., a blue truck influencing the color or motion of a red car), ensuring localized feature refinement.

\textbf{T2V/V2T Masking.}
For each entity’s local prompt $T_i$, we restrict its attention to video tokens within its dynamic bounding box regions $V_i$. This ensures that textual attributes (e.g., color, motion) influence only the intended spatial-temporal regions.

\textbf{V2V Masking.}
Video tokens are partitioned by entity bounding boxes. Tokens within the same entity region attend freely, while tokens from different entities interact only if their bounding boxes overlap spatially (modeling occlusions or interactions). When entities $E_i$ and $E_j$ occupy overlapping regions, their tokens $V_{E_i}$ and $V_{E_j}$ are merged into a single interaction class, allowing localized co-modeling.

\subsection{Entity-Consistency Constraints}

While attention masks offer spatial control over entity generation, their application inherently disrupts the native video generation process, often compromising output quality. For example, temporal inconsistencies, such as flickering artifacts, may arise due to insufficient frame-to-frame coherence. To mitigate this, we utilize dynamically generated bounding boxes derived from the earlier processing stages to control the generation of the later frames. During denoising, tokens within an entity’s first-frame bounding box are designated as reference features. These features are then injected into corresponding tokens within the same entity’s dynamic bounding boxes across subsequent frames. This dynamic feature propagation mechanism strengthens inter-frame feature alignment for each entity, directly counteracting the temporal instability introduced by attention masks.

The synergy between dynamic feature propagation and attention masks establishes dual control over spatial and temporal dimensions. Attention masks enforce localized token interactions, ensuring precise spatial adherence to prompts, while feature propagation maintains entity-specific consistency across frames. Together, these mechanisms balance spatial accuracy with temporal continuity, resulting in higher-quality, artifact-reduced video generation.

\section{Experiments} \label{sec:experiments}

\subsection{Experimental Details}

\textbf{Implementation Details.}
We build upon the diffusers for video generation and develop our model using the DiT-based model, CogVideoX-5B \cite{cogvideox}. Additionally, we utilize DeepSeek-R1 \cite{guo2025deepseek} as the LLM planner. All videos are generated using an NVIDIA H20 GPU, with each clip containing 49 frames at 8 frames per second (fps), resulting in 6.125-second durations.

\textbf{Evaluation Dataset and Metrics.}
We evaluate our framework on T2V-CompBench \cite{t2vcompbench}, a benchmark tailored for evaluating compositional text-to-video generation.
The benchmark consists of seven distinct task categories: Consistent attribute binding, Dynamic attribute binding, Spatial relationships, Motion binding, Action binding, Object interactions, and Generative numeracy, each containing 200 text prompts. These tasks comprehensively assess a model's ability to handle the attributes, quantities, actions, interactions, and spatio-temporal dynamics of compositional text.
For performance evaluation, T2V-CompBench uses 13 metrics derived from various evaluation models, with per-task metric selection based on the highest correlation to human judgment scores, ensuring alignment with human perceptual quality.
Specifically, Grid-LLaVA \cite{t2vcompbench} is used to evaluate Consistent attribute binding, Action binding, and Object interactions, while D-LLaVA assesses Dynamic attribute binding. Grounding-DINO \cite{groundingdino} is used for Spatial relationships and Generative numeracy, and DOT \cite{dot} evaluates Motion binding. Each video receives a score ranging from 0 to 1, with a higher value indicating better alignment with the compositional text description. The final task score for each task is the average of the 200 generated videos' scores, ensuring a robust and standardized evaluation framework for compositional text-to-video generation models.

\textbf{Competitors.}
We benchmark our method against state-of-the-art models across three categories: diffusion U-Net-based models (e.g., AnimateDiff \cite{animatediff}, VideoCrafter2 \cite{videocrafter2}), DiT-based frameworks (e.g., Latte \cite{latte}, Mochi \cite{mochi}), and commercial solutions (e.g., Pika-1.0 \cite{pika}, Kling-1.0 \cite{kling}).

\definecolor{mygray}{gray}{0.9}
\definecolor{myblue}{RGB}{218,232,252}

\begin{table*}[ht]
\centering
    \caption{Evaluation results on T2V-CompBench. The results of all the competitors are from \cite{t2vcompbench}.}
    \resizebox{0.99\linewidth}{!}{
    \setlength{\tabcolsep}{2.4mm}
    \begin{tabular}{l|ccccccc}
        \toprule
        \textbf{Model} & \textbf{Consist-attr} & \textbf{Dynamic-attr} & \textbf{Spatial} & \textbf{Motion} & \textbf{Action} & \textbf{Interaction} & \textbf{Numeracy} \\
        \midrule
        \textbf{Diffusion Unet based} \\
        ModelScope \cite{modelscope} & 51.48\% & 1.61\% & 41.18\% & 24.08\% & 36.39\% & 46.13\% & 19.86\% \\
        ZeroScope \cite{zeroscope} & 40.11\% & 0.91\% & 42.87\% & 24.54\% & 36.61\% & 41.96\% & 24.08\% \\
        LVD \cite{lvd} & 54.39\% & 1.71\% & 54.05\% & 24.57\% & 38.02\% & 45.02\% & 20.08\% \\
        AnimateDiff \cite{animatediff} & 43.25\% & 0.97\% & 39.20\% & 22.27\% & 28.44\% & 39.70\% & 17.67\% \\
        Show-1 \cite{show1} & 56.70\% & 1.15\% & 45.44\% & 22.91\% & 38.81\% & 62.44\% & 30.86\% \\
        VideoCrafter2 \cite{videocrafter2} & 61.82\% & 1.03\% & 48.38\% & 22.59\% & 50.30\% & 63.65\% & 33.30\% \\
        VideoTetris \cite{videotetris} & 62.11\% & 1.04\% & 48.32\% & 22.49\% & 49.39\% & 65.78\% & 34.67\% \\
        Vico \cite{vico} & 58.87\% & 1.07\% & 49.74\% & 22.19\% & 51.11\% & 59.57\% & 32.30\% \\
        T2V-Turbo-V2 \cite{t2vturbov2} & 67.23\% & 1.27\% & 50.25\% & 25.56\% & 60.87\% & 64.39\% & 32.61\% \\
        \midrule
        \textbf{DiT based} \\
        Latte \cite{latte} & 47.13\% & 0.80\% & 43.40\% & 21.55\% & 41.46\% & 41.46\% & 23.20\% \\
        Open-Sora 1.1 \cite{opensora} & 54.14\% & 1.09\% & 54.06\% & 22.61\% & 50.37\% & 55.65\% & 22.59\% \\
        Open-Sora 1.2 \cite{opensora} & 56.39\% & 1.89\% & 50.63\% & 24.68\% & 48.33\% & 50.39\% & 37.19\% \\
        Open-Sora-Plan v1.0.0 \cite{opensoraplan} & 42.46\% & 0.86\% & 45.20\% & 21.48\% & 40.09\% & 41.50\% & 23.31\% \\
        Open-Sora-Plan v1.3.0 \cite{opensoraplan} & 60.76\% & 1.19\% & 51.62\% & 23.77\% & 45.24\% & 44.83\% & 29.52\% \\
        Mochi \cite{mochi} & 59.73\% & 2.46\% & 54.80\% & 23.34\% & 47.59\% & 53.81\% & 27.18\% \\
        \midrule
        \rowcolor{mygray}
        \textbf{Commercial} & & & & & & & \\
        \rowcolor{mygray}
        Pika-1.0 \cite{pika} & 55.36\% & 1.28\% & 46.50\% & 22.34\% & 42.50\% & 51.98\% & 38.70\% \\
      \rowcolor{mygray}
        Gen-2 \cite{gen2} & 57.95\% & 1.09\% & 51.26\% & 21.73\% & 44.13\% & 61.44\% & 30.39\% \\
        \rowcolor{mygray}
        Gen-3 \cite{gen3} & 59.80\% & 6.87\% & 51.94\% & 27.54\% & 52.33\% & 59.06\% & 23.06\% \\
        \rowcolor{mygray}
        Dreamina 1.2 \cite{dreamina} & 69.13\% & 0.51\% & 57.73\% & 23.61\% & 59.24\% & 68.24\% & 43.80\% \\
        \rowcolor{mygray}
        Kling-1.0 \cite{kling} & 69.31\% & 0.98\% & 56.90\% & 25.62\% & 57.87\% & 71.28\% & 44.13\% \\
        \midrule
        CogVideoX-5B \cite{cogvideox} & 61.64\% & 2.19\% & 51.72\% & 26.58\% & 53.33\% & 60.69\% & 37.06\% \\
        \rowcolor{myblue} + Ours & 86.96\% & 2.21\% & 61.10\% & 27.12\% & 73.21\%  & 65.36\% & 39.69\% \\
        \bottomrule
    \end{tabular}
    }
    \label{tab:mainresult}
\end{table*}

\subsection{Experimental Results}
\textbf{Quantitative Comparisons.} \cref{tab:mainresult} presents the quantitative evaluation results of our method and the existing competitors under seven metrics. From the results, we observe that our method demonstrates superior performance across various metrics, showcasing its effectiveness in generating high-quality videos from text prompts. Specifically, DyST-XL achieves the highest scores in Consist-attr, Spatial, Motion, and Action metrics. For example, the Consist-attr score of DyST-XL is 86.96\%, significantly higher than the second-best open-source model T2V-Turbo-V2, which scored 67.23\%. Remarkably, it even surpasses the commercial model Kling-1.0, which has a score of 69.31\%. This indicates a substantial improvement over the baseline model CogVideoX-5B, which scored 25.32\%, demonstrating DyST-XL's enhanced attribute-binding capability. For the challenging Dynamic-attr metric, DyST-XL also exhibits the base CogVideoX model. In terms of the Spatial metric, DyST-XL achieves the best accuracy at 61.10\%, outperforming both the Kling-1.0 commercial model's 56.90\% and the DiT-based model Mochi's 54.80\%. When evaluating the Action metric, DyST-XL attains an accuracy of 73.21\%, which is significantly higher than the second-best model T2V-Turbo-V2 (60.87\%) and the baseline model CogVideoX-5B (53.33\%). Additionally, in the Interaction metric, DyST-XL records an accuracy of 65.36\%, outperforming the baseline model's 60.69\%. Furthermore, for the Numeracy metric, DyST-XL also achieves a higher accuracy of 2.63\% compared to the baseline model, underscoring its proficiency in handling numerical data in video generation. Overall, these results underline DyST-XL's comprehensive advancements across multiple evaluation criteria.

\begin{figure}[t]
    \centering
    \includegraphics[width=\linewidth]{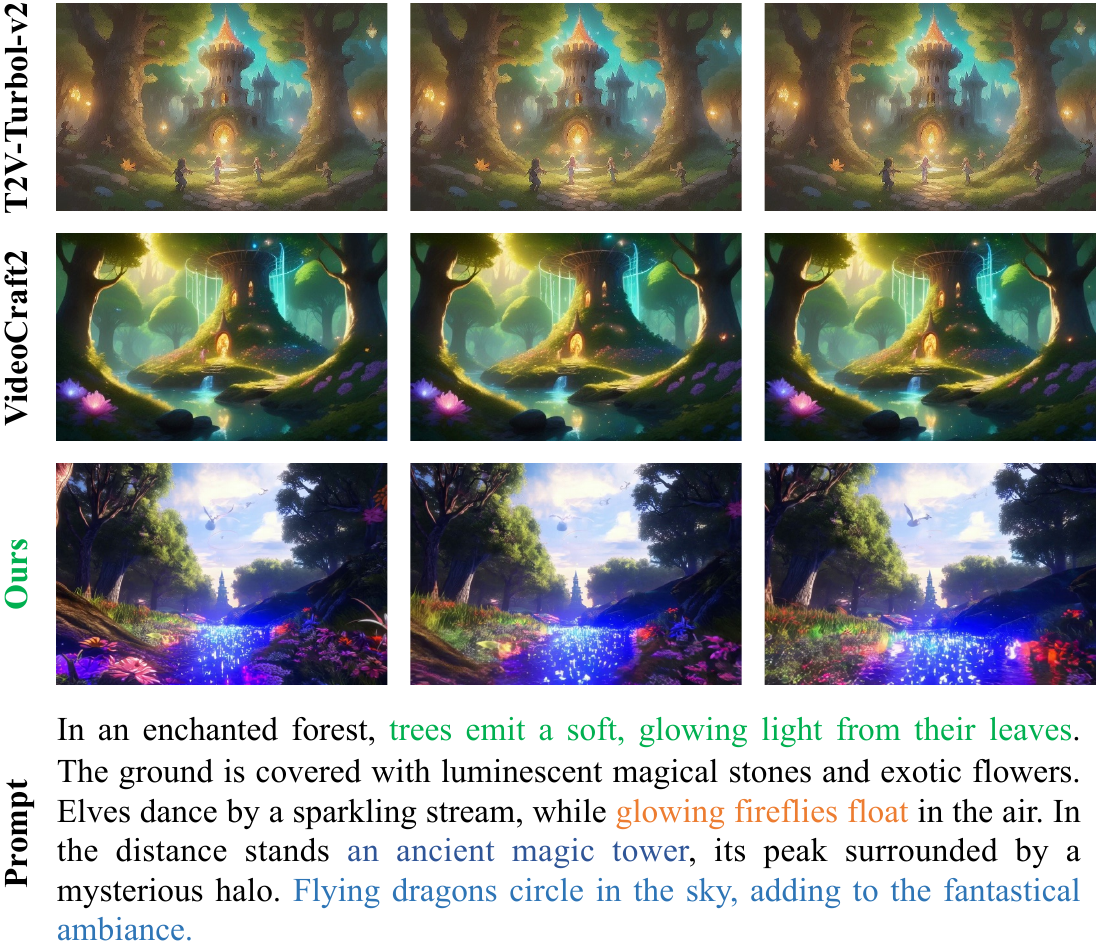}
    \caption{Qualitative comparison on complex prompts.}
    \label{fig: complex prompt results}
\end{figure}

\begin{figure*}[!t]
    \centering
    \includegraphics[width=\linewidth]{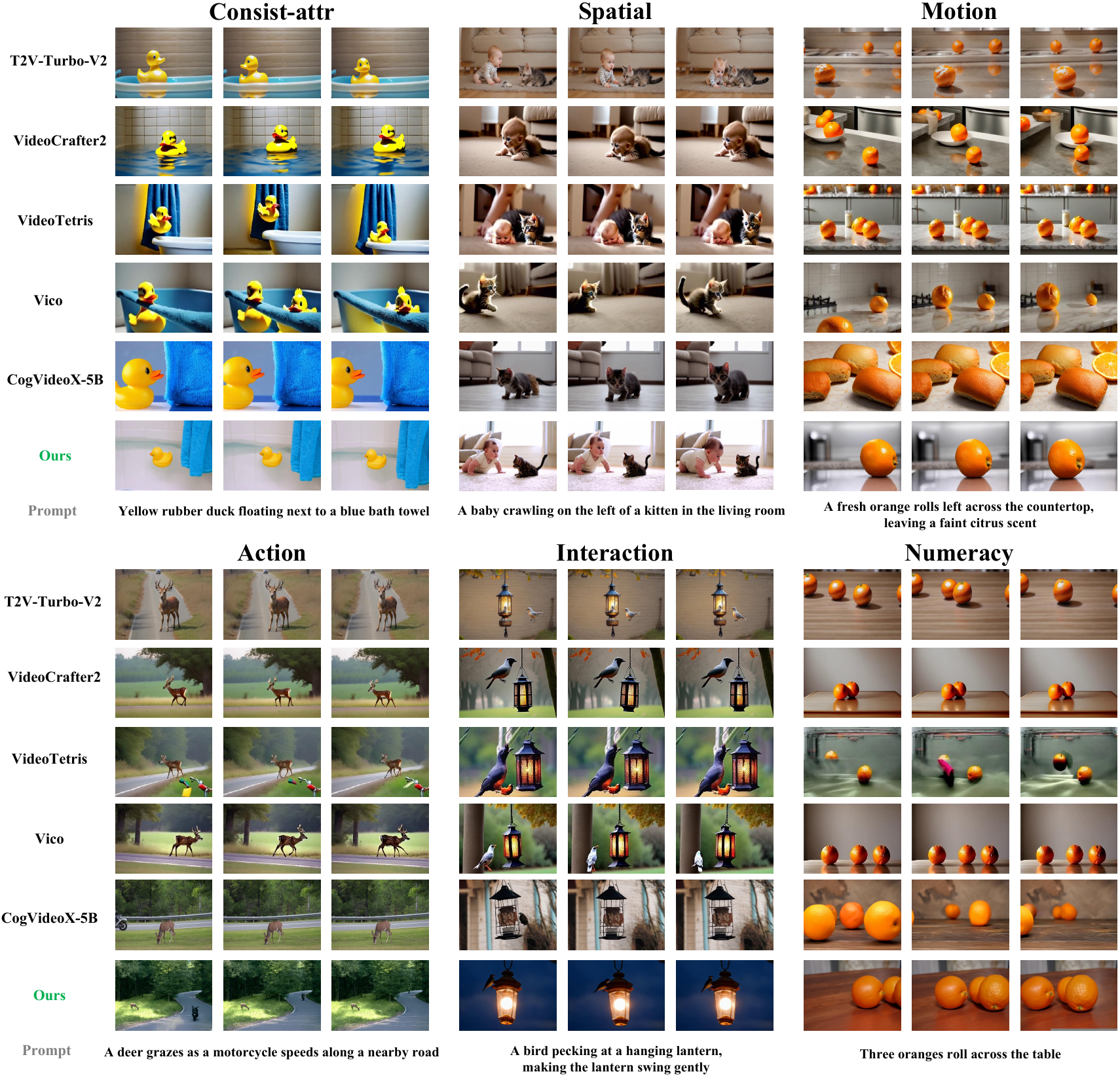}
    \caption{Qualitative visualization comparison with current text-to-video and compositional generation models. The prompts are selected from the T2V-CompBench. (Please zoom in for details.)}
    \label{fig:main-vis}
\end{figure*}

\begin{figure*}[!t]
    \centering
    \begin{overpic}[width=1.0\linewidth]{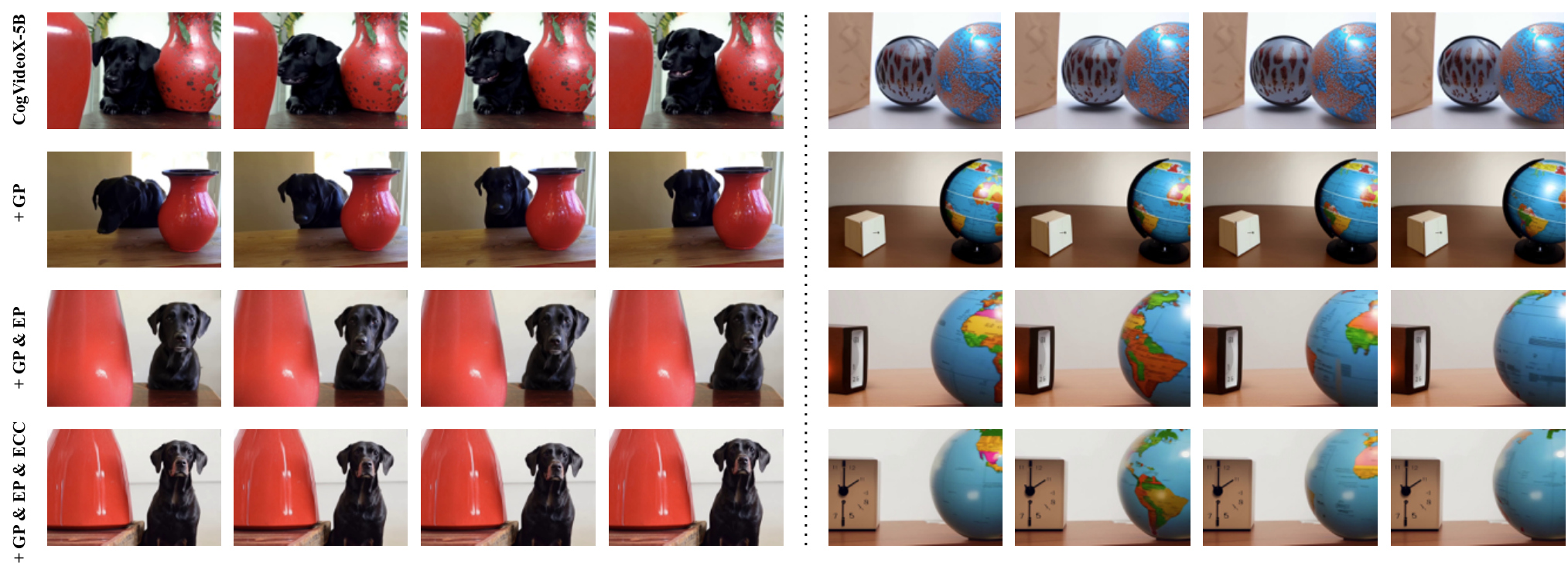}
    \put(10,-2){ {\small \textbf{Prompt:} A black dog sits beside a red vase on a table
}}
    \put(60,-2){ {\small \textbf{Prompt:} Spherical globe rotating next to a cube clock
}}
    \end{overpic}
    \vspace{0.1cm}
    \caption{Qualitative ablation studies on different modules.}
    \label{fig:ab-results}
\end{figure*}

\subsubsection{Qualitative Comparisons}
\cref{fig:main-vis} qualitatively compares the generative performance of {VideoCrafter2} \cite{videocrafter2}, {VideoTetris} \cite{videotetris}, {Vico} \cite{vico}, {CogVideoX-5B} \cite{cogvideox}, and our proposed {DyST-XL} in terms of consistency of attributes, spatial relationships, motion, actions, interactive dynamics, and numerical understanding. Specifically, we analyze representative prompts selected from the T2V-CompBench dataset. Both {VideoCrafter2} and {CogVideoX-5B} struggle to maintain consistent object properties and spatial relationships, leading to inaccurate visual representations and compromised temporal coherence. For instance, when describing a scene with the prompt ``yellow rubber duck floating on a blue bath towel," {CogVideoX-5B} fails to accurately capture the duck’s position in the pool due to the ambiguity introduced by the word ``floating." In contrast, {DyST-XL} demonstrates a clear and accurate representation of the continuous attribute relationship, effectively depicting the duck in the pool. Furthermore, {DyST-XL} excels in the representation of motion and action, displaying improved temporal consistency. These qualitative findings highlight that {DyST-XL} outperforms other methods, including the baseline CogVideoX-5B and more sophisticated generative models such as {Vico} and {VideoTetris}. Additionally, our framework is training-free, which further underscores its superiority in this domain.

To further validate robustness, \cref{fig: complex prompt results} illustrates {DyST-XL}'s performance on intricate prompts involving multi-object interactions, diverse attribute combinations, and semantically rich textual descriptions. The results demonstrate consistent adherence to complex scene specifications, confirming the framework's capacity to parse and execute lengthy, compositionally dense prompts with high fidelity.

\subsection{Ablation Studies}
\textbf{Designed Modules.}
In the ablation study, we evaluate the impact of various modules on the performance of the baseline CogVideoX-5B across three key attributes: Consist-attr, Spatial, and Numeracy, as shown in \cref{tab:ab-modules}. The baseline model achieves scores of 61.64\%, 51.72\%, and 37.06\% for these attributes, respectively. Each module progressively enhances the model's performance. The introduction of the Global Prompt (GP) module improves all attributes, increasing the scores to 65.79\%, 53.38\%, and 37.91\%. GP provides a global context, improving the model's ability to maintain consistent attribute representations. The addition of the Entity-Level Prompt (EP) further refines performance, particularly in Consist-attr, which reaches 76.40\%. EP optimizes attention mechanisms, enhancing spatial alignment and consistency. Finally, incorporating the Entity-Consistency Constraints (ECC) strategy leads to the highest performance, with Consist-attr increasing to 86.96\%, Spatial improving to 61.10\%, and Numeracy rising to 39.69\%. These results demonstrate that each component contributes to performance improvement, with the most significant gains observed in the Consist-attr and Spatial metrics.

\begin{table}[t]
    \centering
    \caption{Impacts (\%) of different modules. GP, EP, and ECC demonstrate Global Prompt, Entity-level Prompt, and Entity-Consistency Constraint, respectively.}
    \resizebox{\linewidth}{!}{
        \begin{tabular}{l|ccc}
        \toprule
            \textbf{Method} & \textbf{Consist-attr} & \textbf{Spatial} & \textbf{Numeracy} \\
        \midrule
            CogVideoX-5B \cite{cogvideox} & 61.64 & 51.72 & 37.06 \\
        \midrule
            + GP & 65.79 & 53.38 & 37.91 \\
            + GP \& EP & 76.40 & 57.93 & 38.21 \\
            + GP \& EP \& ECC & \textbf{86.96} & \textbf{61.10} & \textbf{39.69} \\
        \bottomrule
        \end{tabular}
    }
    \label{tab:ab-modules}
\end{table}

\textbf{Visualization Results.}
\cref{fig:ab-results} further demonstrates the ablation results of our designed modules. The baseline (Row 1), CogVideoX-5B, reveals limitations in accurately associating object attributes and orientation—evidenced by its erroneous generation of a Spherical Clock instead of the intended Spherical globe or Cube clock. With the addition of a global semantic prompt (Row 2), semantic alignment improves significantly, though object-level inaccuracies persist (e.g., the absence of a structurally coherent clock). Incorporating entity-level prompts (Row 3) achieves substantial gains in object fidelity, enabling precise control over individual entities. Finally, integrating our entity-consistency constraint (Row 4) bolsters temporal coherence through feature propagation, producing the most consistent and semantically faithful results.

\begin{figure}
    \centering
    \includegraphics[width=\linewidth]{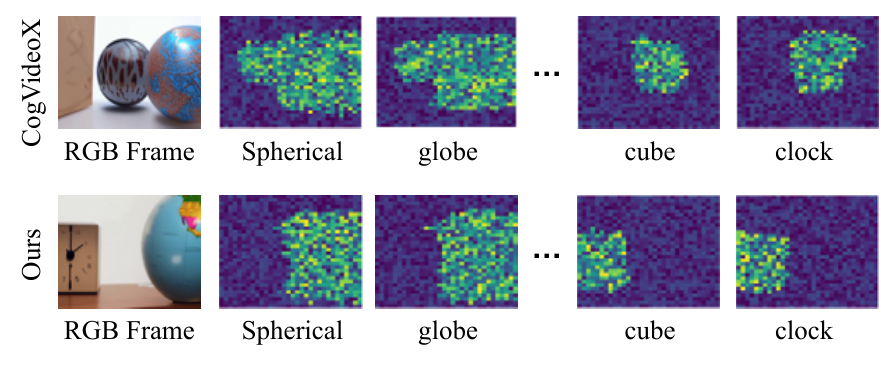}
    \caption{Attention heatmap visualization for CogVideoX-5B and our DyST-XL using the prompt ``Spherical globe rotating next to a cube clock".}
    \label{fig: ab-featuremap}
\end{figure}

\textbf{Feature Map.} \cref{fig: ab-featuremap} illustrates the attention heatmap between the textual prompts and video features. Specifically, we select the first frame for visualization. Compared with the baseline model CogVideoX-5B, our method demonstrates more precise focus allocation, accurately highlighting the spatial relationships between geometric elements as described in the text prompt. 

More details and experimental results are provided in \textbf{Supplementary Materials}.

\section{Conclusion} \label{sec: conclusion}

Our work addresses the persistent challenges in compositional text-to-video generation by introducing DyST-XL, a training-free framework that augments off-the-shelf diffusion models with spatiotemporal control mechanisms. Through systematic experimentation and analysis, we demonstrate that the integration of physics-aware layout planning, global-local prompt attention control, and entity-centric feature propagation enables the synthesis of complex dynamic scenes with high fidelity and coherence. Beyond quantitative improvements on benchmarks, qualitative results reveal DyST-XL’s ability to handle intricate prompts involving multi-object interactions, where existing methods often fail to preserve object roles or physical causality. This advancement positions training-free frameworks as viable alternatives for resource-constrained scenarios, particularly in applications requiring rapid iteration.

\bibliographystyle{ACM-Reference-Format}
\bibliography{sample-base}


\begin{thebibliography}{44}


\ifx \showCODEN    \undefined \def \showCODEN     #1{\unskip}     \fi
\ifx \showISBNx    \undefined \def \showISBNx     #1{\unskip}     \fi
\ifx \showISBNxiii \undefined \def \showISBNxiii  #1{\unskip}     \fi
\ifx \showISSN     \undefined \def \showISSN      #1{\unskip}     \fi
\ifx \showLCCN     \undefined \def \showLCCN      #1{\unskip}     \fi
\ifx \shownote     \undefined \def \shownote      #1{#1}          \fi
\ifx \showarticletitle \undefined \def \showarticletitle #1{#1}   \fi
\ifx \showURL      \undefined \def \showURL       {\relax}        \fi
\providecommand\bibfield[2]{#2}
\providecommand\bibinfo[2]{#2}
\providecommand\natexlab[1]{#1}
\providecommand\showeprint[2][]{arXiv:#2}

\bibitem[Achiam et~al\mbox{.}(2023)]%
        {gpt-4}
\bibfield{author}{\bibinfo{person}{Josh Achiam}, \bibinfo{person}{Steven Adler}, \bibinfo{person}{Sandhini Agarwal}, \bibinfo{person}{Lama Ahmad}, \bibinfo{person}{Ilge Akkaya}, \bibinfo{person}{Florencia~Leoni Aleman}, \bibinfo{person}{Diogo Almeida}, \bibinfo{person}{Janko Altenschmidt}, \bibinfo{person}{Sam Altman}, \bibinfo{person}{Shyamal Anadkat}, {et~al\mbox{.}}} \bibinfo{year}{2023}\natexlab{}.
\newblock \showarticletitle{Gpt-4 technical report}.
\newblock \bibinfo{journal}{\emph{arXiv preprint arXiv:2303.08774}} (\bibinfo{year}{2023}).
\newblock


\bibitem[Capcut(2024)]%
        {dreamina}
\bibfield{author}{\bibinfo{person}{Capcut}.} \bibinfo{year}{2024}\natexlab{}.
\newblock \bibinfo{title}{Dreamina}.
\newblock \bibinfo{howpublished}{\url{https://dreamina.capcut.com/ai-tool/home}}.
\newblock


\bibitem[Chen et~al\mbox{.}(2024)]%
        {videocrafter2}
\bibfield{author}{\bibinfo{person}{Haoxin Chen}, \bibinfo{person}{Yong Zhang}, \bibinfo{person}{Xiaodong Cun}, \bibinfo{person}{Menghan Xia}, \bibinfo{person}{Xintao Wang}, \bibinfo{person}{Chao Weng}, {and} \bibinfo{person}{Ying Shan}.} \bibinfo{year}{2024}\natexlab{}.
\newblock \showarticletitle{Videocrafter2: Overcoming data limitations for high-quality video diffusion models}. In \bibinfo{booktitle}{\emph{CVPR}}. \bibinfo{pages}{7310--7320}.
\newblock


\bibitem[Esser et~al\mbox{.}(2024)]%
        {sd3}
\bibfield{author}{\bibinfo{person}{Patrick Esser}, \bibinfo{person}{Sumith Kulal}, \bibinfo{person}{Andreas Blattmann}, \bibinfo{person}{Rahim Entezari}, \bibinfo{person}{Jonas M{\"u}ller}, \bibinfo{person}{Harry Saini}, \bibinfo{person}{Yam Levi}, \bibinfo{person}{Dominik Lorenz}, \bibinfo{person}{Axel Sauer}, \bibinfo{person}{Frederic Boesel}, {et~al\mbox{.}}} \bibinfo{year}{2024}\natexlab{}.
\newblock \showarticletitle{Scaling rectified flow transformers for high-resolution image synthesis}. In \bibinfo{booktitle}{\emph{ICML}}.
\newblock


\bibitem[Guo et~al\mbox{.}(2025)]%
        {guo2025deepseek}
\bibfield{author}{\bibinfo{person}{Daya Guo}, \bibinfo{person}{Dejian Yang}, \bibinfo{person}{Haowei Zhang}, \bibinfo{person}{Junxiao Song}, \bibinfo{person}{Ruoyu Zhang}, \bibinfo{person}{Runxin Xu}, \bibinfo{person}{Qihao Zhu}, \bibinfo{person}{Shirong Ma}, \bibinfo{person}{Peiyi Wang}, \bibinfo{person}{Xiao Bi}, {et~al\mbox{.}}} \bibinfo{year}{2025}\natexlab{}.
\newblock \showarticletitle{Deepseek-r1: Incentivizing reasoning capability in llms via reinforcement learning}.
\newblock \bibinfo{journal}{\emph{arXiv preprint arXiv:2501.12948}} (\bibinfo{year}{2025}).
\newblock


\bibitem[Guo et~al\mbox{.}(2023)]%
        {animatediff}
\bibfield{author}{\bibinfo{person}{Yuwei Guo}, \bibinfo{person}{Ceyuan Yang}, \bibinfo{person}{Anyi Rao}, \bibinfo{person}{Zhengyang Liang}, \bibinfo{person}{Yaohui Wang}, \bibinfo{person}{Yu Qiao}, \bibinfo{person}{Maneesh Agrawala}, \bibinfo{person}{Dahua Lin}, {and} \bibinfo{person}{Bo Dai}.} \bibinfo{year}{2023}\natexlab{}.
\newblock \showarticletitle{Animatediff: Animate your personalized text-to-image diffusion models without specific tuning}.
\newblock \bibinfo{journal}{\emph{arXiv preprint arXiv:2307.04725}} (\bibinfo{year}{2023}).
\newblock


\bibitem[He et~al\mbox{.}(2022)]%
        {LVDM}
\bibfield{author}{\bibinfo{person}{Yingqing He}, \bibinfo{person}{Tianyu Yang}, \bibinfo{person}{Yong Zhang}, \bibinfo{person}{Ying Shan}, {and} \bibinfo{person}{Qifeng Chen}.} \bibinfo{year}{2022}\natexlab{}.
\newblock \showarticletitle{Latent video diffusion models for high-fidelity long video generation}.
\newblock \bibinfo{journal}{\emph{arXiv preprint arXiv:2211.13221}} (\bibinfo{year}{2022}).
\newblock


\bibitem[Ho et~al\mbox{.}(2020)]%
        {ho2020ddpm}
\bibfield{author}{\bibinfo{person}{Jonathan Ho}, \bibinfo{person}{Ajay Jain}, {and} \bibinfo{person}{Pieter Abbeel}.} \bibinfo{year}{2020}\natexlab{}.
\newblock \showarticletitle{Denoising diffusion probabilistic models}.
\newblock \bibinfo{journal}{\emph{NeurIPS}}  \bibinfo{volume}{33}, \bibinfo{pages}{6840--6851}.
\newblock


\bibitem[Khachatryan et~al\mbox{.}(2023)]%
        {text2videozero}
\bibfield{author}{\bibinfo{person}{Levon Khachatryan}, \bibinfo{person}{Andranik Movsisyan}, \bibinfo{person}{Vahram Tadevosyan}, \bibinfo{person}{Roberto Henschel}, \bibinfo{person}{Zhangyang Wang}, \bibinfo{person}{Shant Navasardyan}, {and} \bibinfo{person}{Humphrey Shi}.} \bibinfo{year}{2023}\natexlab{}.
\newblock \showarticletitle{Text2video-zero: Text-to-image diffusion models are zero-shot video generators}. In \bibinfo{booktitle}{\emph{ICCV}}. \bibinfo{pages}{15954--15964}.
\newblock


\bibitem[Kuaishou(2024)]%
        {kling}
\bibfield{author}{\bibinfo{person}{Kuaishou}.} \bibinfo{year}{2024}\natexlab{}.
\newblock \bibinfo{title}{Kling}.
\newblock \bibinfo{howpublished}{\url{https://kling.kuaishou.com}}.
\newblock


\bibitem[Labs(2024)]%
        {flux}
\bibfield{author}{\bibinfo{person}{Black~Forest Labs}.} \bibinfo{year}{2024}\natexlab{}.
\newblock \bibinfo{title}{FLUX}.
\newblock \bibinfo{howpublished}{\url{https://github.com/black-forest-labs/flux}}.
\newblock


\bibitem[Le~Moing et~al\mbox{.}(2024)]%
        {dot}
\bibfield{author}{\bibinfo{person}{Guillaume Le~Moing}, \bibinfo{person}{Jean Ponce}, {and} \bibinfo{person}{Cordelia Schmid}.} \bibinfo{year}{2024}\natexlab{}.
\newblock \showarticletitle{Dense optical tracking: Connecting the dots}. In \bibinfo{booktitle}{\emph{CVPR}}. \bibinfo{pages}{19187--19197}.
\newblock


\bibitem[Lee et~al\mbox{.}(2024)]%
        {videorepair}
\bibfield{author}{\bibinfo{person}{Daeun Lee}, \bibinfo{person}{Jaehong Yoon}, \bibinfo{person}{Jaemin Cho}, {and} \bibinfo{person}{Mohit Bansal}.} \bibinfo{year}{2024}\natexlab{}.
\newblock \showarticletitle{VideoRepair: Improving Text-to-Video Generation via Misalignment Evaluation and Localized Refinement}.
\newblock \bibinfo{journal}{\emph{arXiv preprint arXiv:2411.15115}} (\bibinfo{year}{2024}).
\newblock


\bibitem[Li et~al\mbox{.}(2024)]%
        {t2vturbov2}
\bibfield{author}{\bibinfo{person}{Jiachen Li}, \bibinfo{person}{Qian Long}, \bibinfo{person}{Jian Zheng}, \bibinfo{person}{Xiaofeng Gao}, \bibinfo{person}{Robinson Piramuthu}, \bibinfo{person}{Wenhu Chen}, {and} \bibinfo{person}{William~Yang Wang}.} \bibinfo{year}{2024}\natexlab{}.
\newblock \showarticletitle{T2v-turbo-v2: Enhancing video generation model post-training through data, reward, and conditional guidance design}.
\newblock \bibinfo{journal}{\emph{arXiv preprint arXiv:2410.05677}} (\bibinfo{year}{2024}).
\newblock


\bibitem[Li et~al\mbox{.}(2023)]%
        {gligen}
\bibfield{author}{\bibinfo{person}{Yuheng Li}, \bibinfo{person}{Haotian Liu}, \bibinfo{person}{Qingyang Wu}, \bibinfo{person}{Fangzhou Mu}, \bibinfo{person}{Jianwei Yang}, \bibinfo{person}{Jianfeng Gao}, \bibinfo{person}{Chunyuan Li}, {and} \bibinfo{person}{Yong~Jae Lee}.} \bibinfo{year}{2023}\natexlab{}.
\newblock \showarticletitle{Gligen: Open-set grounded text-to-image generation}.
\newblock  (\bibinfo{year}{2023}), \bibinfo{pages}{22511--22521}.
\newblock


\bibitem[Lian et~al\mbox{.}(2023)]%
        {lvd}
\bibfield{author}{\bibinfo{person}{Long Lian}, \bibinfo{person}{Baifeng Shi}, \bibinfo{person}{Adam Yala}, \bibinfo{person}{Trevor Darrell}, {and} \bibinfo{person}{Boyi Li}.} \bibinfo{year}{2023}\natexlab{}.
\newblock \showarticletitle{Llm-grounded video diffusion models}.
\newblock \bibinfo{journal}{\emph{arXiv preprint arXiv:2309.17444}} (\bibinfo{year}{2023}).
\newblock


\bibitem[Lin et~al\mbox{.}(2024)]%
        {opensoraplan}
\bibfield{author}{\bibinfo{person}{Bin Lin}, \bibinfo{person}{Yunyang Ge}, \bibinfo{person}{Xinhua Cheng}, \bibinfo{person}{Zongjian Li}, \bibinfo{person}{Bin Zhu}, \bibinfo{person}{Shaodong Wang}, \bibinfo{person}{Xianyi He}, \bibinfo{person}{Yang Ye}, \bibinfo{person}{Shenghai Yuan}, \bibinfo{person}{Liuhan Chen}, {et~al\mbox{.}}} \bibinfo{year}{2024}\natexlab{}.
\newblock \showarticletitle{Open-sora plan: Open-source large video generation model}.
\newblock \bibinfo{journal}{\emph{arXiv preprint arXiv:2412.00131}} (\bibinfo{year}{2024}).
\newblock


\bibitem[Liu et~al\mbox{.}(2023)]%
        {groundingdino}
\bibfield{author}{\bibinfo{person}{Shilong Liu}, \bibinfo{person}{Zhaoyang Zeng}, \bibinfo{person}{Tianhe Ren}, \bibinfo{person}{Feng Li}, \bibinfo{person}{Hao Zhang}, \bibinfo{person}{Jie Yang}, \bibinfo{person}{Chunyuan Li}, \bibinfo{person}{Jianwei Yang}, \bibinfo{person}{Hang Su}, \bibinfo{person}{Jun Zhu}, {et~al\mbox{.}}} \bibinfo{year}{2023}\natexlab{}.
\newblock \showarticletitle{Grounding dino: Marrying dino with grounded pre-training for open-set object detection}.
\newblock \bibinfo{journal}{\emph{arXiv preprint arXiv:2303.05499}} (\bibinfo{year}{2023}).
\newblock


\bibitem[Ma et~al\mbox{.}(2025)]%
        {stepvideot2v}
\bibfield{author}{\bibinfo{person}{Guoqing Ma}, \bibinfo{person}{Haoyang Huang}, \bibinfo{person}{Kun Yan}, \bibinfo{person}{Liangyu Chen}, \bibinfo{person}{Nan Duan}, \bibinfo{person}{Shengming Yin}, \bibinfo{person}{Changyi Wan}, \bibinfo{person}{Ranchen Ming}, \bibinfo{person}{Xiaoniu Song}, \bibinfo{person}{Xing Chen}, {et~al\mbox{.}}} \bibinfo{year}{2025}\natexlab{}.
\newblock \showarticletitle{Step-video-t2v technical report: The practice, challenges, and future of video foundation model}.
\newblock \bibinfo{journal}{\emph{arXiv preprint arXiv:2502.10248}} (\bibinfo{year}{2025}).
\newblock


\bibitem[Ma et~al\mbox{.}(2024)]%
        {latte}
\bibfield{author}{\bibinfo{person}{Xin Ma}, \bibinfo{person}{Yaohui Wang}, \bibinfo{person}{Gengyun Jia}, \bibinfo{person}{Xinyuan Chen}, \bibinfo{person}{Ziwei Liu}, \bibinfo{person}{Yuan-Fang Li}, \bibinfo{person}{Cunjian Chen}, {and} \bibinfo{person}{Yu Qiao}.} \bibinfo{year}{2024}\natexlab{}.
\newblock \showarticletitle{Latte: Latent diffusion transformer for video generation}.
\newblock \bibinfo{journal}{\emph{arXiv preprint arXiv:2401.03048}} (\bibinfo{year}{2024}).
\newblock


\bibitem[Peebles and Xie(2023)]%
        {DiT}
\bibfield{author}{\bibinfo{person}{William Peebles} {and} \bibinfo{person}{Saining Xie}.} \bibinfo{year}{2023}\natexlab{}.
\newblock \showarticletitle{Scalable Diffusion Models with Transformers}. In \bibinfo{booktitle}{\emph{ICCV}}. \bibinfo{pages}{4172--4182}.
\newblock


\bibitem[Pika(2024)]%
        {pika}
\bibfield{author}{\bibinfo{person}{Pika}.} \bibinfo{year}{2024}\natexlab{}.
\newblock \bibinfo{title}{Pika}.
\newblock \bibinfo{howpublished}{\url{https://www.pika.art}}.
\newblock


\bibitem[Rombach et~al\mbox{.}(2022)]%
        {stable_diffusion}
\bibfield{author}{\bibinfo{person}{Robin Rombach}, \bibinfo{person}{Andreas Blattmann}, \bibinfo{person}{Dominik Lorenz}, \bibinfo{person}{Patrick Esser}, {and} \bibinfo{person}{Bj{\"{o}}rn Ommer}.} \bibinfo{year}{2022}\natexlab{}.
\newblock \showarticletitle{High-Resolution Image Synthesis with Latent Diffusion Models}. In \bibinfo{booktitle}{\emph{CVPR}}. \bibinfo{pages}{10674--10685}.
\newblock


\bibitem[Ronneberger et~al\mbox{.}(2015)]%
        {ronneberger2015u}
\bibfield{author}{\bibinfo{person}{Olaf Ronneberger}, \bibinfo{person}{Philipp Fischer}, {and} \bibinfo{person}{Thomas Brox}.} \bibinfo{year}{2015}\natexlab{}.
\newblock \showarticletitle{U-net: Convolutional networks for biomedical image segmentation}. In \bibinfo{booktitle}{\emph{MICCAI}}. \bibinfo{pages}{234--241}.
\newblock


\bibitem[Runway(2024a)]%
        {gen2}
\bibfield{author}{\bibinfo{person}{Runway}.} \bibinfo{year}{2024}\natexlab{a}.
\newblock \bibinfo{title}{Gen-2: Generate novel videos with text, images or video clips}.
\newblock \bibinfo{howpublished}{\url{https://research.runwayml.com/gen2}}.
\newblock


\bibitem[Runway(2024b)]%
        {gen3}
\bibfield{author}{\bibinfo{person}{Runway}.} \bibinfo{year}{2024}\natexlab{b}.
\newblock \bibinfo{title}{Introducing gen-3 alpha: A new frontier for video generation}.
\newblock \bibinfo{howpublished}{\url{https://runwayml.com/blog/introducing-gen-3-alpha}}.
\newblock


\bibitem[Song et~al\mbox{.}(2020)]%
        {ddim}
\bibfield{author}{\bibinfo{person}{Jiaming Song}, \bibinfo{person}{Chenlin Meng}, {and} \bibinfo{person}{Stefano Ermon}.} \bibinfo{year}{2020}\natexlab{}.
\newblock \showarticletitle{Denoising Diffusion Implicit Models}.
\newblock \bibinfo{journal}{\emph{CoRR}}  \bibinfo{volume}{abs/2010.02502} (\bibinfo{year}{2020}).
\newblock
\showeprint[arXiv]{2010.02502}
\urldef\tempurl%
\url{https://arxiv.org/abs/2010.02502}
\showURL{%
\tempurl}


\bibitem[Sterling(2023)]%
        {zeroscope}
\bibfield{author}{\bibinfo{person}{Spencer Sterling}.} \bibinfo{year}{2023}\natexlab{}.
\newblock \bibinfo{title}{Zeroscope}.
\newblock \bibinfo{howpublished}{\url{https://huggingface.co/cerspense/zeroscope_v2_576w}}.
\newblock


\bibitem[Sun et~al\mbox{.}(2024)]%
        {t2vcompbench}
\bibfield{author}{\bibinfo{person}{Kaiyue Sun}, \bibinfo{person}{Kaiyi Huang}, \bibinfo{person}{Xian Liu}, \bibinfo{person}{Yue Wu}, \bibinfo{person}{Zihan Xu}, \bibinfo{person}{Zhenguo Li}, {and} \bibinfo{person}{Xihui Liu}.} \bibinfo{year}{2024}\natexlab{}.
\newblock \showarticletitle{T2v-compbench: A comprehensive benchmark for compositional text-to-video generation}.
\newblock \bibinfo{journal}{\emph{arXiv preprint arXiv:2407.14505}} (\bibinfo{year}{2024}).
\newblock


\bibitem[Team(2024)]%
        {mochi}
\bibfield{author}{\bibinfo{person}{Genmo Team}.} \bibinfo{year}{2024}\natexlab{}.
\newblock \bibinfo{title}{Mochi 1}.
\newblock \bibinfo{howpublished}{\url{https://github.com/genmoai/models}}.
\newblock


\bibitem[Tian et~al\mbox{.}(2024)]%
        {videotetris}
\bibfield{author}{\bibinfo{person}{Ye Tian}, \bibinfo{person}{Ling Yang}, \bibinfo{person}{Haotian Yang}, \bibinfo{person}{Yuan Gao}, \bibinfo{person}{Yufan Deng}, \bibinfo{person}{Xintao Wang}, \bibinfo{person}{Zhaochen Yu}, \bibinfo{person}{Xin Tao}, \bibinfo{person}{Pengfei Wan}, \bibinfo{person}{Di ZHANG}, {et~al\mbox{.}}} \bibinfo{year}{2024}\natexlab{}.
\newblock \showarticletitle{Videotetris: Towards compositional text-to-video generation}.
\newblock \bibinfo{journal}{\emph{NeurIPS}}  \bibinfo{volume}{37} (\bibinfo{year}{2024}), \bibinfo{pages}{29489--29513}.
\newblock


\bibitem[Wang et~al\mbox{.}(2023)]%
        {modelscope}
\bibfield{author}{\bibinfo{person}{Jiuniu Wang}, \bibinfo{person}{Hangjie Yuan}, \bibinfo{person}{Dayou Chen}, \bibinfo{person}{Yingya Zhang}, \bibinfo{person}{Xiang Wang}, {and} \bibinfo{person}{Shiwei Zhang}.} \bibinfo{year}{2023}\natexlab{}.
\newblock \showarticletitle{Modelscope text-to-video technical report}.
\newblock \bibinfo{journal}{\emph{arXiv preprint arXiv:2308.06571}} (\bibinfo{year}{2023}).
\newblock


\bibitem[Wang et~al\mbox{.}(2024a)]%
        {InstanceDiffusion}
\bibfield{author}{\bibinfo{person}{Xudong Wang}, \bibinfo{person}{Trevor Darrell}, \bibinfo{person}{Sai~Saketh Rambhatla}, \bibinfo{person}{Rohit Girdhar}, {and} \bibinfo{person}{Ishan Misra}.} \bibinfo{year}{2024}\natexlab{a}.
\newblock \showarticletitle{Instancediffusion: Instance-level control for image generation}. In \bibinfo{booktitle}{\emph{CVPR}}. \bibinfo{pages}{6232--6242}.
\newblock


\bibitem[Wang et~al\mbox{.}(2024b)]%
        {dreamrunner}
\bibfield{author}{\bibinfo{person}{Zun Wang}, \bibinfo{person}{Jialu Li}, \bibinfo{person}{Han Lin}, \bibinfo{person}{Jaehong Yoon}, {and} \bibinfo{person}{Mohit Bansal}.} \bibinfo{year}{2024}\natexlab{b}.
\newblock \showarticletitle{DreamRunner: Fine-Grained Storytelling Video Generation with Retrieval-Augmented Motion Adaptation}.
\newblock \bibinfo{journal}{\emph{arXiv preprint arXiv:2411.16657}} (\bibinfo{year}{2024}).
\newblock


\bibitem[Wu et~al\mbox{.}(2023)]%
        {tuneavideo}
\bibfield{author}{\bibinfo{person}{Jay~Zhangjie Wu}, \bibinfo{person}{Yixiao Ge}, \bibinfo{person}{Xintao Wang}, \bibinfo{person}{Stan~Weixian Lei}, \bibinfo{person}{Yuchao Gu}, \bibinfo{person}{Yufei Shi}, \bibinfo{person}{Wynne Hsu}, \bibinfo{person}{Ying Shan}, \bibinfo{person}{Xiaohu Qie}, {and} \bibinfo{person}{Mike~Zheng Shou}.} \bibinfo{year}{2023}\natexlab{}.
\newblock \showarticletitle{Tune-a-video: One-shot tuning of image diffusion models for text-to-video generation}. In \bibinfo{booktitle}{\emph{ICCV}}. \bibinfo{pages}{7623--7633}.
\newblock


\bibitem[Yang et~al\mbox{.}(2024b)]%
        {RPG}
\bibfield{author}{\bibinfo{person}{Ling Yang}, \bibinfo{person}{Zhaochen Yu}, \bibinfo{person}{Chenlin Meng}, \bibinfo{person}{Minkai Xu}, \bibinfo{person}{Stefano Ermon}, {and} \bibinfo{person}{Bin Cui}.} \bibinfo{year}{2024}\natexlab{b}.
\newblock \showarticletitle{Mastering text-to-image diffusion: Recaptioning, planning, and generating with multimodal llms}.
\newblock \bibinfo{journal}{\emph{arXiv preprint arXiv:2401.11708}} (\bibinfo{year}{2024}).
\newblock


\bibitem[Yang and Wang(2024)]%
        {vico}
\bibfield{author}{\bibinfo{person}{Xingyi Yang} {and} \bibinfo{person}{Xinchao Wang}.} \bibinfo{year}{2024}\natexlab{}.
\newblock \showarticletitle{Compositional video generation as flow equalization}.
\newblock \bibinfo{journal}{\emph{arXiv preprint arXiv:2407.06182}} (\bibinfo{year}{2024}).
\newblock


\bibitem[Yang et~al\mbox{.}(2024a)]%
        {cogvideox}
\bibfield{author}{\bibinfo{person}{Zhuoyi Yang}, \bibinfo{person}{Jiayan Teng}, \bibinfo{person}{Wendi Zheng}, \bibinfo{person}{Ming Ding}, \bibinfo{person}{Shiyu Huang}, \bibinfo{person}{Jiazheng Xu}, \bibinfo{person}{Yuanming Yang}, \bibinfo{person}{Wenyi Hong}, \bibinfo{person}{Xiaohan Zhang}, \bibinfo{person}{Guanyu Feng}, {et~al\mbox{.}}} \bibinfo{year}{2024}\natexlab{a}.
\newblock \showarticletitle{Cogvideox: Text-to-video diffusion models with an expert transformer}.
\newblock \bibinfo{journal}{\emph{arXiv preprint arXiv:2408.06072}} (\bibinfo{year}{2024}).
\newblock


\bibitem[Zhang et~al\mbox{.}(2024)]%
        {show1}
\bibfield{author}{\bibinfo{person}{David~Junhao Zhang}, \bibinfo{person}{Jay~Zhangjie Wu}, \bibinfo{person}{Jia-Wei Liu}, \bibinfo{person}{Rui Zhao}, \bibinfo{person}{Lingmin Ran}, \bibinfo{person}{Yuchao Gu}, \bibinfo{person}{Difei Gao}, {and} \bibinfo{person}{Mike~Zheng Shou}.} \bibinfo{year}{2024}\natexlab{}.
\newblock \showarticletitle{Show-1: Marrying pixel and latent diffusion models for text-to-video generation}.
\newblock \bibinfo{journal}{\emph{IJCV}} (\bibinfo{year}{2024}), \bibinfo{pages}{1--15}.
\newblock


\bibitem[Zhang et~al\mbox{.}(2023)]%
        {controlnet}
\bibfield{author}{\bibinfo{person}{Lvmin Zhang}, \bibinfo{person}{Anyi Rao}, {and} \bibinfo{person}{Maneesh Agrawala}.} \bibinfo{year}{2023}\natexlab{}.
\newblock \showarticletitle{Adding Conditional Control to Text-to-Image Diffusion Models}. In \bibinfo{booktitle}{\emph{ICCV}}. \bibinfo{pages}{3813--3824}.
\newblock


\bibitem[Zhang et~al\mbox{.}(2025)]%
        {regionalpromptflux}
\bibfield{author}{\bibinfo{person}{Xinyu Zhang}, \bibinfo{person}{Zicheng Duan}, \bibinfo{person}{Dong Gong}, {and} \bibinfo{person}{Lingqiao Liu}.} \bibinfo{year}{2025}\natexlab{}.
\newblock \showarticletitle{Training-Free Motion-Guided Video Generation with Enhanced Temporal Consistency Using Motion Consistency Loss}.
\newblock \bibinfo{journal}{\emph{arXiv preprint arXiv:2501.07563}} (\bibinfo{year}{2025}).
\newblock


\bibitem[Zheng et~al\mbox{.}(2024)]%
        {opensora}
\bibfield{author}{\bibinfo{person}{Zangwei Zheng}, \bibinfo{person}{Xiangyu Peng}, \bibinfo{person}{Tianji Yang}, \bibinfo{person}{Chenhui Shen}, \bibinfo{person}{Shenggui Li}, \bibinfo{person}{Hongxin Liu}, \bibinfo{person}{Yukun Zhou}, \bibinfo{person}{Tianyi Li}, {and} \bibinfo{person}{Yang You}.} \bibinfo{year}{2024}\natexlab{}.
\newblock \showarticletitle{Open-sora: Democratizing efficient video production for all}.
\newblock \bibinfo{journal}{\emph{arXiv preprint arXiv:2412.20404}} (\bibinfo{year}{2024}).
\newblock


\bibitem[Zhou et~al\mbox{.}(2024)]%
        {MIGC}
\bibfield{author}{\bibinfo{person}{Dewei Zhou}, \bibinfo{person}{You Li}, \bibinfo{person}{Fan Ma}, \bibinfo{person}{Xiaoting Zhang}, {and} \bibinfo{person}{Yi Yang}.} \bibinfo{year}{2024}\natexlab{}.
\newblock \showarticletitle{Migc: Multi-instance generation controller for text-to-image synthesis}. In \bibinfo{booktitle}{\emph{CVPR}}. \bibinfo{pages}{6818--6828}.
\newblock


\bibitem[Zhu et~al\mbox{.}(2023)]%
        {zhu2023minigpt}
\bibfield{author}{\bibinfo{person}{Deyao Zhu}, \bibinfo{person}{Jun Chen}, \bibinfo{person}{Xiaoqian Shen}, \bibinfo{person}{Xiang Li}, {and} \bibinfo{person}{Mohamed Elhoseiny}.} \bibinfo{year}{2023}\natexlab{}.
\newblock \showarticletitle{Minigpt-4: Enhancing vision-language understanding with advanced large language models}.
\newblock \bibinfo{journal}{\emph{arXiv preprint arXiv:2304.10592}} (\bibinfo{year}{2023}).
\newblock


\end{thebibliography}











\end{document}